\documentclass{article}

\usepackage[final]{neurips_2020}
\usepackage[utf8]{inputenc} 
\usepackage[T1]{fontenc}    
\usepackage{hyperref}       
\usepackage{url}            
\usepackage{booktabs}       
\usepackage{amsfonts}       
\usepackage{nicefrac}       
\usepackage{microtype}      
\usepackage{graphicx}
\usepackage{multirow}
\usepackage[title]{appendix}
\usepackage[ruled,vlined]{algorithm2e}
\usepackage{xr-hyper}
\usepackage{hyperref}
\hypersetup{
	colorlinks=true,
	linkcolor=blue,
	filecolor=magenta,      
	urlcolor=cyan,
}
\setcitestyle{square,sort,comma,numbers}

\newtheorem{definition}{Definition}
\title{Reinforced Molecular Optimization with Neighborhood-Controlled Grammars}

\author{
	Chencheng Xu,\textsuperscript{\rm 1,2}
	Qiao Liu,\textsuperscript{\rm 1,3}
	Minlie Huang,\textsuperscript{\rm 1,2$*$}
	Tao Jiang\textsuperscript{\rm 4,1,2}\thanks{Minlie Huang and Tao Jiang are the co-corresponding authors.}\\
	\textsuperscript{\rm 1}BNRIST, Tsinghua University, Beijing 100084, China\\
	\textsuperscript{\rm 2}Department of Computer Science and Technology, Tsinghua University, Beijing 100084, China\\
	\textsuperscript{\rm 3}Department of Automation, Tsinghua University, Beijing 100084, China\\
	\textsuperscript{\rm 4}Department of Computer Science and Engineering, UCR, CA 92521, USA\\
	\texttt{\{xucc18, liu-q16\}@mails.tsinghua.edu.cn}\\
	\texttt{ aihuang@tsinghua.edu.cn, jiang@cs.ucr.edu}
}

\begin{document}
	
	\maketitle
	
	\begin{abstract}
		A major challenge in the pharmaceutical industry is to design novel molecules with specific desired properties, especially when the property evaluation is costly. Here, we propose MNCE-RL, a graph convolutional policy network for molecular optimization with molecular neighborhood-controlled embedding grammars through reinforcement learning. We extend the original neighborhood-controlled embedding grammars to make them applicable to molecular graph generation and design an efficient algorithm to infer grammatical production rules from given molecules. The use of grammars guarantees the validity of the generated molecular structures. By transforming molecular graphs to parse trees with the inferred grammars, the molecular structure generation task is modeled as a Markov decision process where a policy gradient strategy is utilized. In a series of experiments, we demonstrate that our approach achieves state-of-the-art performance in a diverse range of molecular optimization tasks and exhibits significant superiority in optimizing molecular properties with a limited number of property evaluations.
	\end{abstract}
	
	\section{Introduction}
	Traditional drug discovery relies on the development and exploration by expert chemists and pharmacologists, which is time-consuming due to the large chemical structure space \cite{walters2018virtual}. Effective methods for collecting chemical structures with desired properties will significantly reduce the number of candidates for wet-lab experiments and thus accelerate the development of novel drugs.
	
	Recently, several methods have been proposed to solve the molecular optimization problem within the deep learning framework \cite{jin2018junction, you2018graph,kajino2019molecular,kusner2017grammar,segler2018generating, gomez2018automatic}. The major challenges for molecular optimization mainly lie in generating valid molecular structures and efficiently exploring the vast chemical structure space. Although several methods, including \cite{weininger1988smiles,kusner2017grammar,jin2018junction,you2018graph}, have been proposed to solve the first challenge, they either involve complex network architectures or struggle to optimize properties due to the choices of molecular representations \cite{jin2018junction,kusner2017grammar}. The second challenge is addressed by Bayesian optimization (BO) \cite{jin2018junction,kusner2017grammar,movckus1975bayesian} and reinforcement learning (RL) \cite{you2018graph}. However, few of these methods considered the high cost to evaluate molecular properties in real-world applications \cite{kajino2019molecular}. In fact, for most chemical and biological properties, such as antibacterial, anticancer and teratogenicity, there are no known explicit functions to directly interpret a chemical structure as a corresponding numerical property score. Hence, time-consuming wet-lab experiments or simulations are typically required to evaluate these properties of molecules, resulting in a limited number of molecules with validated properties. Therefore, generating molecules with desired properties using a small number of property evaluations as well as a small number of molecules with known properties is critical.
	
\begin{figure}
		\centering
		\includegraphics[width=\textwidth]{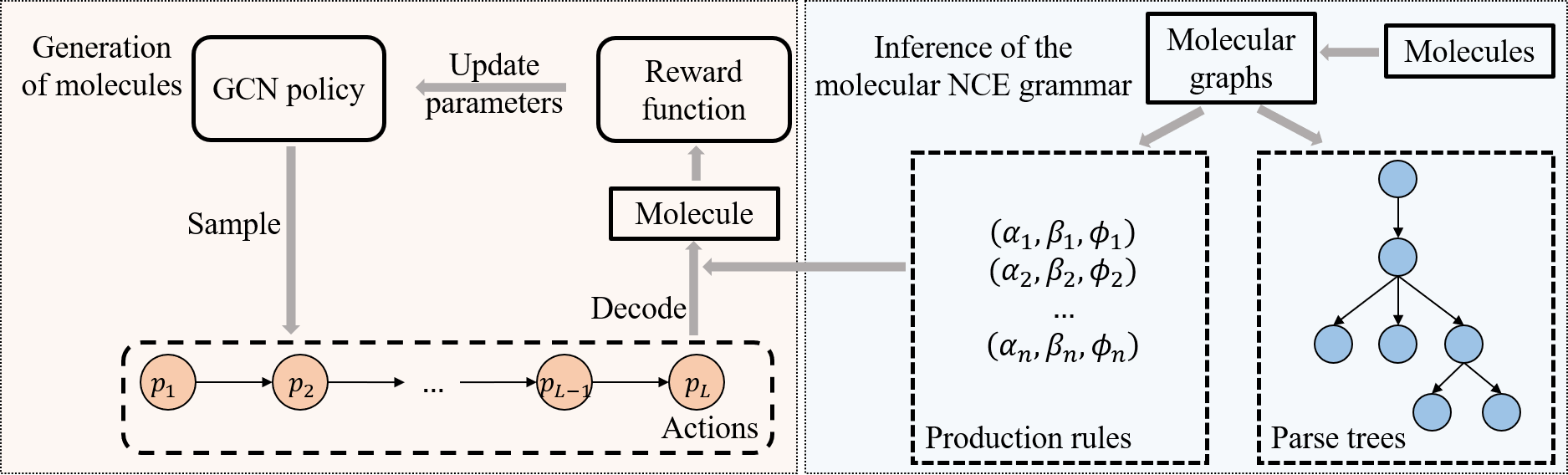}
		\caption{Illustration of our framework. We first infer a molecular NCE grammar by representing molecules as molecular graphs, parsing the graphs using neighborhood-controlled rules, and extracting the production rules. In the generation process, a GCN-based policy network samples a sequence of productions from the action space and obtains rewards from the reward function. The reward function measures the specific property of the molecule decoded from the generated action sequence.}
		\label{procedure}
	\end{figure}

	To tackle these challenges, we propose MNCE-RL, an RL-based framework using the proposed molecular neighborhood-controlled embedding grammars and a graph convolutional network (GCN). The molecular neighborhood-controlled embedding graph grammars are extended from neighborhood-controlled embedding (NCE) grammars \cite{fahmy1992survey,janssens1982graph}, which are a type of sequential context-free graph grammars. As shown in Figure \ref{procedure}, a molecular NCE grammar can be inferred from the input molecular graphs so that each molecule can be represented as a parse tree. In the generation process, an RL agent generates a sequence of production rules, and receives a reward from the environment, which measures the specific property of the generated molecule, that can be used to update the GCN policy network. Our proposed molecular NCE grammars guarantee the chemical validity and the RL agent can efficiently explore the vast chemical structure space.

	Our major contributions include 1) a novel molecular NCE grammar and an efficient algorithm to infer production rules from given molecules, where the grammar provides a way to simplify the generation of valid molecules; 2) a novel GCN architecture updating both node and edge features to compute feature vectors for nodes in molecular graphs, where the update of edge features in the GCN makes it possible to capture subtle physical differences between bonds with the same labels and thus lead to better node features for policy decision making; 3) the experimental results show that MNCE-RL significantly outperforms state-of-the-art methods in molecular optimization and has a high potential to be useful in drug discovery.
	
	\section{Related work}
	Early methods \cite{segler2018generating,gomez2018automatic,dai2018syntax,guimaraes2017objective} represent molecules as SMILES strings \cite{weininger1988smiles}, where the generation of a molecule is modeled as a Markov decision process (MDP) and recurrent neural networks are used to generate the SMILES string. Compared to the graph representation, the SMILES representation is quite brittle as a small change in the string may lead to a completely different molecule, which makes it hard to optimize molecular properties \cite{kajino2019molecular}. {Winter \textit{et al}. \cite{winter2019efficient} optimize molecular properties in a continuous latent space learned from SMILES strings to overcome the brittleness of the SMILES representation.} Li \textit{et al}. \cite{li2018learning} first attempt to generate molecules with the graph representation and achieves promising results in generating novel and realistic molecules, but their method cannot guarantee the validity of the generated molecules. To reduce the ratio of invalid molecules, Jin \textit{et al}. \cite{jin2018junction} (JT-VAE) proposes to represent molecules with junction trees where each node in the tree represents a cluster of atoms and optimize properties in the latent space of the variational autoencoder (VAE) by BO. Although the chemical validity constraints are intrinsically satisfied by predefined connections in clusters, uncertainty in combining the generated clusters limits the model's ability to optimize molecular properties. You \textit{et al}. \cite{you2018graph} (GCPN) try to generate molecular graphs by iteratively adding atoms and edges using a graph convolutional policy network and guarantee the chemical validity by the imposition of certain chemical constraints on generated structures. Due to its complex model architecture, GCPN requires a large number of iterations in training, which limits its applications in situations when property evaluation is costly. Kajino \cite{kajino2019molecular} (MHG-VAE) is the first to apply graph grammars to the molecular optimization problem. With a simple VAE architecture, MHG-VAE shows superiority in molecular optimization with a limited number of property evaluations. However, the performance of MHG-VAE is still far from being satisfactory perhaps due to the choice of the grammars and indirect optimization in a latent space.
	
	\section{Methods}
	As mentioned in \cite{kajino2019molecular}, molecular optimization can be formulated as follows:
	\begin{equation}
	m^*={\arg\max}_{m\in \mathcal{M}} f(m),
	\end{equation}
	where $\mathcal{M}$ is the set of all valid chemical molecules and $f$ is an evaluation function, which measures some specific property score of molecule $m$. We represent a molecule as a graph $H=(V, E, \sigma, \psi)$ by modeling atoms as nodes and bonds as edges, where $V$ is a finite set of nodes, $E$ is a finite set of edges, $\sigma: V\to \Sigma $ is a node-labeling function, which projects $V$ to the node label set $\Sigma$, and similarly $\psi: E\to \Psi$ is an edge-labeling function that projects $E$ to the edge label set $\Psi$. Following \cite{kajino2019molecular}, we use the {Kekul\'e} structure of molecules and include the chirality tag in node labels.
	
	Using the proposed molecular NCE grammars, the generation of a novel molecule is interpreted as the generation of a parse tree, where each node in the tree represents a production rule. Furthermore, by traversing the parse trees in preorder, the molecular optimization problem is interpreted as the generation of an optimal production sequence, {\it i.e.} 
	\begin{equation}
	Prod^*={\arg\max}_{Prod\in \mathcal{P}}f\circ{Dec_\mathcal{P}}(Prod),
	\end{equation}
	
	where $\mathcal{P}$ is the set of all valid sequences of production rules and $Dec_\mathcal{P}:\mathcal{P}\to \mathcal{M} $ is the decoding function that transforms a production sequence into a molecule. The problem can be cast as an MDP and solved in the RL framework, where a GCN is used for node feature aggregation. Given an intermediate production sequence $Prod_t$ generated at time step $t$, due to the constraints of the molecular NCE grammar, the next production rule can only be selected from a subset of the production rules. We denote a production rule to be legal for $Prod_t$ if it satisfying the grammatical constraints.
	
	\subsection{Problem formulation as reinforcement learning}
	As aforementioned, the generation of sequences of production rules can be formulated as a sequential decision problem. Hence, we present the design of state representation, action space, and reward function as follows.
	
	{\bf State.} We denote the state $s_t$ at time step $t$ as the intermediate sequence $Prod_t=p_1p_2...p_{t-1}$, from which a graph $H_t$ can be decoded and the non-terminal node $v_{t}$ to be rewritten at time step $t+1$ is determined. Note that at the first step, $Prod_1$ is an empty sequence and $H_1$ has only one node $v_1$ with the starting symbol. 
	
	{\bf Action.} The action space is a set of the legal production rules for $Prod_t$. In time step $t$, the policy $\pi_\theta(a_t|s_t)$ samples a production rule from the action space, where
	\begin{equation}
	\pi_\theta(a_t|s_t)=softmax(\mathbf{F}_{\theta'}(H_t)_{v_t}\mathbf{W}+\mathbf{b}),
	\end{equation}
	in which $\mathbf{F}$ is a GCN described in section \ref{aggnet}, $\theta'$ is the parameter set of $\mathbf{F}$, and $\theta=\{\mathbf{W}, \mathbf{b}\}\cup \theta'$.  $\mathbf{F}_{\theta'}(H_t)$ is the computed node feature matrix of $H_t$ and $\mathbf{F}_{\theta'}(H_t)_{v_t}$ is the row corresponding to the node $v_t$. The intermediate molecular graph is updated with the sampled production rule.
	
	{\bf Reward.} As the generation process may take too many steps to converge, we set a threshold $T_{max}$ and force the generation process to stop when the number of steps exceeds $T_{max}$. Assume that the length of the generated sequence is $T-1$. At time step $t<T$, a small constant reward $r_\epsilon$ is assigned and at time step $T$, if there is no non-terminal node in $H_{T}$, a task-specific reward function assigns a reward based on $f\circ Dec_{\mathcal{P}}(H_{T})$. Otherwise, a constant non-positive reward $r_{incomp}$ is assigned.
	
	\subsection{Definition of molecular NCE grammars}
	An NCE graph grammar proposed by Janssens et al. \cite{janssens1982graph} is a system $G=(\Sigma, \Delta_\Sigma, P)$, where $\Sigma$ is the set of node labels, and $\Delta_\Sigma\subset\Sigma$ is the terminal alphabet and $P$ is the set of production rules. A production rule is in the form of $p=(\alpha, \beta, \phi)$, where $\alpha$, $\beta$ are connected graphs. $\alpha$ is called the left-hand side (LHS) of $p$, $\beta$ is called the right-hand side (RHS), and $\phi:V_\alpha \times V_\beta\times \Sigma\to \{0,1\}$ is the embedding function. Directly applying NCE grammars to molecular graphs suffers from the following issues: 1) A molecular graph is both node-labeled and edge-labeled, while the NCE grammars are defined only on node-labeled graphs. 2) The connections between the neighbors of $V_\alpha$ and nodes in $V_\beta$ are not specified, which may cause valency invalidity in a molecular graph. 3) The number of production rules may explode, decreasing the generalization ability of the grammars. To extend NCE grammars to molecular graphs, we define molecular NCE grammars as follows.
	\begin{definition}
		A molecular NCE grammar is a system $G=(\Sigma, \Psi, \Delta_\Sigma, \Delta_\Psi, P)$, where $\Sigma$ is the set of node labels, $\Psi$ the set of edge labels, $\Delta_\Sigma=\Sigma\setminus\{x,n_\Sigma, s \}$ the terminal alphabet of nodes, $\Delta_\Psi=\Psi\setminus \{n_\Psi\}$ the terminal alphabet of edges, $s$ the starting symbol, and $n_\Sigma$ and $n_\Psi$ the empty labels for nodes and edges, respectively. Finally, $P$ is the set of production rules. A production rule is in the form of $p=(\alpha, \beta, \phi)$ where:
		\begin{itemize}
			\item $\alpha=(V_\alpha, E_\alpha, \sigma_\alpha, \psi_\alpha, L_\alpha)$ and $\beta=(V_\beta, E_\beta, \sigma_\beta, \psi_\beta, L_\beta)$ are ordered connected graphs, where $L_*$ defines a unique order for edges incident to each vertex in the graph
			\item $V_\alpha=\{X_p \}\cup \mathcal{B}_p$, $E_\alpha=\{X_p\}\times \mathcal{B}_p$, where$X_p$ is a non-terminal node with $\sigma_\alpha(X_p)=x$ and $\mathcal{B}_p$ is a set of nodes with $\forall v\in \mathcal{B}_p$, $\sigma_\alpha(v)=n_\Sigma$
			\item $V_\beta=\mathcal{T}_p\cup \mathcal{N}_p$, $E_\beta\subset (\mathcal{T}_p\times\mathcal{T}_p)\cup (\mathcal{T}_p\times\mathcal{N}_p) $, where $\mathcal{T}_p$ and $\mathcal{N}_p$ are sets of nodes with $\forall v\in \mathcal{N}_p, \sigma_\beta(v)=x$
			\begin{itemize}
				\item if $\|\mathcal{T}_p\|>1$, then $\forall u \in \mathcal{T}_p, \sigma_\beta(u)=n_\Sigma$, $\forall e\in E_\beta, \psi_\beta(e)=n_\Psi$, $p$ is a complex production rule
				\item if $\|\mathcal{T}_p\|=1$, then $\forall u \in \mathcal{T}_p$, $\sigma_\beta(u)\in\Delta_\Sigma$, $\forall e\in E_\beta, \psi_\beta(e)\in\Delta_\Psi$, $p$ is a simple production rule
			\end{itemize} 
			\item $\phi: \mathcal{B}_p\times \mathcal{T}_p\to \Psi\cup \{0 \}$ is the embedding function
		\end{itemize}
	\end{definition}
	
	\begin{figure}
		\centering
		\includegraphics[width=0.9\textwidth]{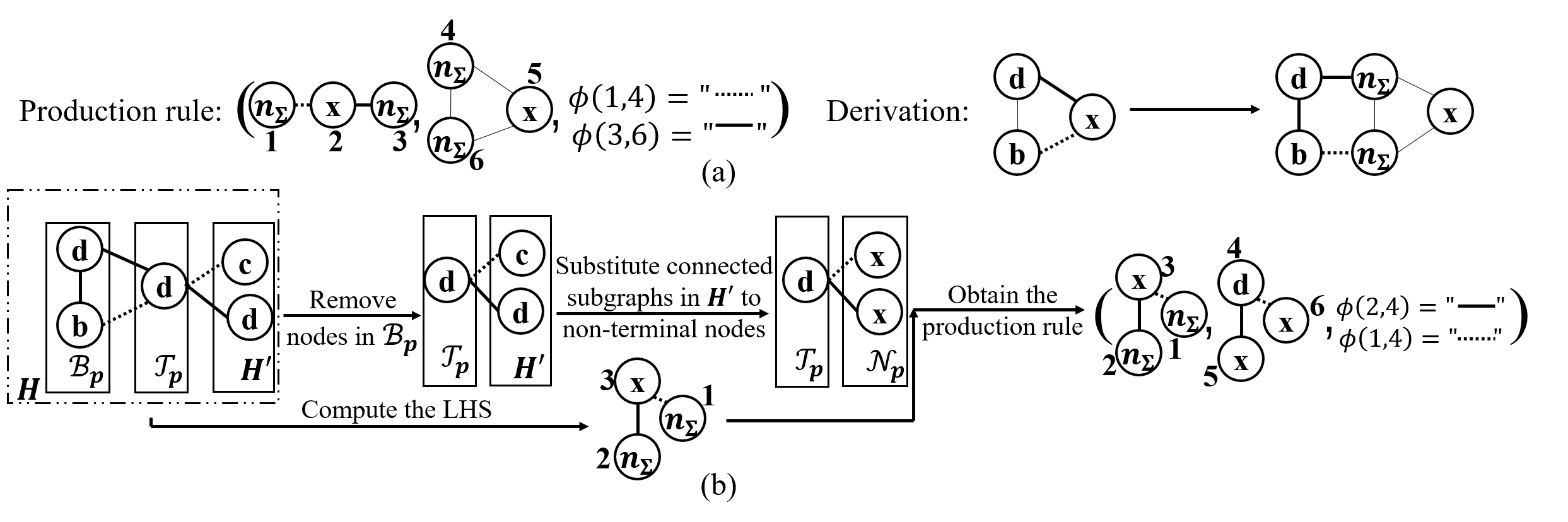}
		\caption{(a) An example production rule and a derivation step. Here, $x$ and $n_\Sigma$ are non-terminal labels. The production rule is in the form of $(\alpha, \beta, \phi)$. In a derivation step, applying a production rule $p$ will replace a non-terminal node $v_t$ (with label $x$) in the intermediate graph $H_t$ with the RHS ($\beta$) of $p$, and the edges between neighbors of $v_t$ and nodes in $V_\beta$ is determined by the embedding function $\phi$. (b) Extraction of a production rule. $\mathcal{B}_p$, $\mathcal{T}_p$ and $\mathcal{N}_p$ are vertex sets. $H'$ is a node-induced subgraph of $H$. The LHS is obtained by representing the nodes in $\mathcal{T}_p$ as a non-terminal node, removing the edges between the nodes in $\mathcal{B}_p$ and labeling the nodes in $\mathcal{B}_p$ as $n_\Sigma$. The RHS is obtained by removing the nodes in $\mathcal{B}_p$ from $H$ and replacing the connected subgraphs in $H'$ by non-terminal nodes.}
		\label{deriveprod}
	\end{figure}
	The first two issues mentioned above are addressed by specifying $\psi_\alpha$, $\psi_\beta$ and $\phi$. To alleviate the third issue, we introduce the empty labels that can be matched arbitrarily, $n_\Sigma$ and $n_\Psi$, in a more general way. The labels of nodes in $\mathcal{B}_p$ are replaced by $n_\Sigma$ and for complex production rules, only the skeletons of $\beta$ are kept. Production rules predefine the edges incident to each vertex and thus the valency validity can be guaranteed intrinsically. To specify the action space at each step, we define legal production rules as follows.
	\begin{definition}
		Let $T_t$ be an intermediate tree. If $T_t$ is an empty tree, the legal production rules for $T_t$ is the set of starting production rules. If $T_t$ is not empty and we need to sample a child production rule for the parent $p_{parent}$ that already has a set of child production rules $P_{sibling}$, then an intermediate graph $H_t$ with a non-terminal node $v_t$ to be rewritten at in the next time step can be decoded from $T_t$. Suppose that the direct neighbors of $v_t$ are $\{ v_{n_1}, v_{n_2}, ..., v_{n_k}\}$ and $L_t$ sorts the edge set $\{(v_t,v_{n_1}), (v_t,v_{n_2}), ..., (v_t,v_{n_k})\}$ in the order in which $v_{n_i}$ are generated, we say that a production rule $p$ matches the context of $v_t$ if and only if the edge-induced subgraph of $H_t$ specified by $\{ (v_t,v_{n_1}), (v_t,v_{n_2}), ..., (v_t,v_{n_k})\}$ and ordered by $L_t$ is isomorphic to the LHS of $p$ \cite{jiang1999optimal}. Then
		
		\begin{itemize}
			\item if $p_{parent}$ is complex and $v_t\in \mathcal{T}_{p_{parent}}$, any production rule having a positive empirical probability $P(p|p_{parent}, P_{sibling})$ and matching the context of $v_t$ is legal for $T_t$
			
			\item otherwise, any production rule matching the context of $v_t$ is legal for $T_t$
		\end{itemize}
		
	\end{definition}
	An example production rule and a derivation step are shown in Figure \ref{deriveprod}. Applying a production rule $p$ to an intermediate graph $H_t$ to rewrite a non-terminal node $v_t$ will replace $v_t$ with the RHS of $p$, and the edges between the direct neighbors of $v_t$ and nodes in the RHS are specified by the embedding function. A formal notion of a derivation step is defined as follows.
	
	\begin{definition}
		\label{derivationstep}
		Let $T_t$ be an intermediate parse tree and a production rule $p=(\alpha, \beta, \phi)$ is legal for $T_t$. An intermediate graph $H_t$ and a non-terminal node $v_t$ can be decoded from $T_t$. A derivation step of applying $p$ to $H_t$ will generate a novel graph $H_{t+1}$ by rewriting the node $v_t$, where
		\begin{itemize}
			\item $V_{H_{t+1}}=V_{H_t}\cup V_\beta\setminus\{v_t\}$
			\item $E_{H_{t+1}}=\{(u,v)|u,v\in V_{H_t}\setminus \{v_t\}\ and\ (u,v)\in E_{H_t} \}\cup E_\beta \cup \{(u,v)|\phi(u,v)\in \Psi \}$
		\end{itemize}
		For a node $u\in V_{H_{t+1}}$ and an edge $(u,v)\in E_{H_{t+1}}$, the labeling functions are:
		\begin{eqnarray*}
			\left\{
			\begin{array}{lr}
				\sigma_{H_{t+1}}(u)=\sigma_{H_t}(u),\ u\in V_{H_t}\\
				\sigma_{H_{t+1}}(u)=\sigma_\beta(u),\ u\in V_\beta\\
			\end{array}
			\right.,
			\left\{
			\begin{array}{lr}
				\psi_{H_{t+1}}\left( (u,v) \right)=\psi_{H_t}\left((u,v)\right),\ (u,v)\in E_{H_t}\\
				\psi_{H_{t+1}}\left( (u,v) \right)=\psi_{\beta}\left((u,v)\right),\ (u,v)\in E_\beta\\
				\psi_{H_{t+1}}\left( (u,v) \right)=\phi(u,v),\ u\in V_{H_t}, v\in V_\beta, \phi(u,v)\in \Psi\\
			\end{array}
			\right.
		\end{eqnarray*}		
	\end{definition}
	{With this definition, by learning production rules from known molecules, any molecule sampled from the inferred grammar is chemically valid. A comparison of our proposed grammars and the MHGs \cite{kajino2019molecular} is shown in Appendix B.}
	
	\subsection{Inference of the molecular NCE grammars}
	The algorithm to parse molecular graphs and infer the production rules is shown in Appendix B. We sort the nodes of $H$ in the depth-first (DF) order, and for a node $v$ with first-hop neighbors $\{v_{n_1}, v_{n_2}, ..., v_{n_k}\}$, the edges $\{(v, v_{n_1}), (v, v_{n_2}), ..., (v, v_{n_k})\}$ are sorted to be consistent with the order of $v_{n_i}$. The graph is parsed in the DF order and the LHS and RHS extracted from $H$ inherit the edge orders. For a simple production rule (Figure \ref{deriveprod}), the LHS is simply obtained by representing nodes in $\mathcal{T}_p$ as a non-terminal node, removing the edges between the nodes in $\mathcal{B}_p$ and labeling nodes in the $\mathcal{B}_p$ as $n_\Sigma$. The embedding function $\phi$ is obtained by recording the edges between the nodes in $\mathcal{T}_p$ and $\mathcal{B}_p$. Denoting the node-induced subgraph of $H$ specified by $V_H\setminus\left(\mathcal{B}_p\cup\mathcal{T}_p\right)$ as $H'$, the RHS is obtained by removing the nodes in $\mathcal{B}_p$ from $H$ and representing each connected subgraph of $H'$ with a non-terminal node. For the complex production rules (Appendix A), the first steps are also computing the LHS, recording the embedding function, removing nodes in $\mathcal{B}_p$, and substitute connected subgraphs in $H'$ into non-terminal nodes. In the final step, as discussed in the prior section, to reduce the number of production rules, we only keep the skeleton of the RHS, and the labels of all nodes in $\mathcal{T}_p$ and the labels of all edges in the RHS are replaced by $n_\Sigma$ and $n_\Psi$. To maintain the information, we introduce an extra production rule for each node in $\mathcal{T}_{p}$. Examples to parse a molecular graph and to sample a molecule from a grammar is shown in Appendix A.
	
	\subsection{Graph convolutional network for node feature aggregation}
	\label{aggnet}
	Graph convolutional networks (GCNs) \cite{gilmer2017neural,Hu2019PretrainingGN,liao2019lanczosnet, li2018adaptive, gao2019graph, kim2019edge} have been widely applied in graph information aggregation. We represent both nodes and edges with feature vectors. In the forward pass, the GCN updates both the node features and the edge features and outputs the computed features for all nodes in the last layer. Assuming that the feature size of the edges is $S_E$, the node features are updated by
	\begin{equation}
	\mathbf{V}^{(l+1)}=AGG(\{Tanh(\mathbf{E}_{(i)}^{(l)}\mathbf{V}^{(l)}\mathbf{W}_{(i)}^{(l+1)}+\mathbf{b}^{(l+1)}_{(i)})+\mathbf{V}^{(l)}| i\in (1, ..., S_E)\}),
	\end{equation}
	
	where $AGG$ is the aggregation function, $\mathbf{V}^{(l)}$ is the node feature matrix in the $l$th layer, $\mathbf{E}^{(l)}_{(i)}$ is the $i$th feature matrix of edges, and $\mathbf{W}_{*}^{*}$ and $\mathbf{b}^{*}_{*}$ are parameters of the network. The edge features are updated in two steps. At the first step, we calculate a vector $\mathbf{e}_{ij}$, which encodes the relationship between the $i$-th node and the $j$-th node using the following formula
	\begin{equation}
	\mathbf{e}_{ij}^{(l+1)}=ReLU(Concat(\mathbf{V}_i^{(l+1)}, \mathbf{V}_j^{(l+1)})\mathbf{W}_{e}^{(l+1)}+\mathbf{b}_{e}^{(l+1)}),
	\end{equation}
	where $\mathbf{V}_i^{(l+1)}$ is the feature vector of the node $i$ in the $(l+1)$th layer. Then, the feature vector $\mathbf{E}_{ij}$ of the edge between the node $i$ and the node $j$ is updated by
	\begin{equation}
	\mathbf{E}_{ij}^{(l+1)}=ReLU(Concat(\mathbf{e}_{ij}^{(l+1)}, \mathbf{E}_{ij}^{(l)})\mathbf{W}_{E}^{(l+1)}+\mathbf{b}_{E}^{(l+1)}).
	\end{equation}
	\subsection{Model training}
	To generate molecules with desired properties, the widely used RL technique, Proximal Policy Optimization \cite{schulman2017proximal} (PPO), is adopted to train the model. The objective function of PPO is 
	\begin{equation}
	L^{CLIP}(\theta)=\hat{E_t}\left[\min(r_t(\theta)\hat{A}_t, clip (r_t(\theta), 1-\epsilon, 1+\epsilon)\hat{A}_t)\right],
	\end{equation}
	where $\epsilon$ is a hyperparameter, $\theta$ is the policy parameter, $\hat{E_t}$ denotes the empirical expectation over timesteps, and $r_t$ is the ratio of the probability under the new and old policies, i.e.
	\begin{equation}
	r_t=\frac{\pi_\theta(a_t|s_t)}{\pi_{\theta_{old}}(a_t|s_t)},
	\end{equation}
	where $\theta_{old}$ is the parameter set of the old policy. $\hat{A_t}$ is the estimated advantage \cite{schulman2015high} at time step $t$. We compute the actor critic $C_\omega (\cdot)$ in $\hat{A_t}$ as
	\begin{equation}
	C_\omega(H_t)=Avg(\mathbf{F}_{\omega'}(H_t))\mathbf{W_C}+\mathbf{b_C},
	\end{equation} 
	where $\mathbf{F}$ is a GCN with the parameter set $\omega'$, $\omega$ is the parameter set of the actor critic and $\omega=\{\mathbf{W_C}, \mathbf{b_C} \}\cup \omega'$. The $Avg$ function computes the average over the node features. To encourage the model to generate graphs with high diversity, an entropy loss \cite{mnih2013playing} is also added to the loss function, and to accelerate convergence, we take all the ground truth molecules as expert trajectories and pre-train the model with these trajectories. {Details of model training and optimizations of hyperparameters are shown in Appendix F}.

	\section{Experiments}
	\subsection{Datasets}
	The ZINC250k molecule dataset \cite{irwin2005zinc}, {GuacaMol package \cite{brown2019guacamol}} and 2,337 drug molecules from \cite{stokes2020deep} are used in our experiments. The ZINC250k dataset contains 250,000 drug-like molecules whose maximum atom number is 38. The work in \cite{stokes2020deep} provides 2,337 drug molecules and their inhibition effects to {\it E.coli} collected from wet-lab experiments. With a threshold of 0.2, 120 of the 2,337 molecules that have a strong {\it E.coli} growth inhibition are defined as the positive set and the remaining molecules are considered as the negative set. {GuacaMol is a comprehensive benchmark package for molecular optimization that provides more than one million molecules and covers not only single-objectives but also constrained and multi-objective optimization tasks}. The validity of generated molecules is checked by RDKit \cite{landrum2013rdkit}. The statistics of the inferred molecular NCE grammars are provided in Appendix C\footnote{Link to code and datasets: \href{https://github.com/Zoesgithub/MNCE-RL}{https://github.com/Zoesgithub/MNCE-RL}}.

	\subsection{Molecular optimization results}
	To demonstrate the ability of MNCE-RL in molecular optimization in different application scenarios, we designed a series of experiments and compared MNCE-RL with the current state-of-the-art methods. Detailed experiment settings of the baseline models \cite{winter2019efficient, kajino2019molecular, jin2018junction, you2018graph} are provided in Appendix D.
	
	{\bf Property optimization with unlimited evaluations and an ablation study.} In this experiment, we assume that the cost of property evaluation is negligible and the number of times to query the molecule properties is unlimited. Penalized logP score and QED score are used to evaluate the performance of models. Here, LogP is an estimation of the octanol-water partition coefficient and penalized logP also accounts for ring size and synthetic accessibility \cite{ertl2009estimation}. QED \cite{bickerton2012quantifying} is a computational score for measuring the drug-likeness of a molecule. To measure the performance of each method, we report the top 3 property scores, the 50th best score, and the average score of the top 50 molecules. The task-specific reward function we used in our approach is a linear projection of the computed penalized logP or QED score. The results are shown in Table \ref{pop} and Appendix G. {To investigate the specific contributions of our proposed grammars and the GCN structure in this experiment, we build a model using the classical GCN \cite{gilmer2017neural,you2018graph} without edge feature updating (MNCE-RL$_{OEU}$). As shown in the tables, MNCE-RL$_{OEU}$ achieves the state-of-the-art performance in optimizing both penalized logP and QED and significantly outperforms GCPN, indicating the effectiveness of our grammars. Moreover, compared with the MHGs, our proposed grammars achieve a higher coverage rate (Appendix C), and thus can represent more molecular structures and explore the chemical space more effectively. The utility of the edge feature updating mechanism is also confirmed by the fact that MNCE-RL outperforms MNCE-RL$_{OEU}$ significantly in optimizing penalized logP.}
	
	\begin{table}[h]
		\setlength\tabcolsep{1.5pt} 
		\caption{Results on property optimizations with unlimited property evalutions}
		\label{pop}
		\centering
		\begin{tabular*}{1\textwidth}{ccccccccccccc}
			\toprule
			\multirow{3}{*}{Method} &\multicolumn{6}{c}{Penalized logP}& \multicolumn{6}{c}{QED}  \\
			\cmidrule(lr){2-7} \cmidrule(lr){8-13}
			& $1^{st}$ &$2^{nd}$ &$3^{rd}$&$50^{th}$&\vtop{\hbox{\strut Top 50}\hbox{\strut\ \ Avg.}} &Validity&$1^{st}$ &$2^{nd}$ &$3^{rd}$&$50^{th}$&\vtop{\hbox{\strut Top 50}\hbox{\strut\ \ Avg.}}&Validity\\
			\cmidrule(lr){1-1}\cmidrule(lr){2-7} \cmidrule(lr){8-13}
			JT-VAE&5.30&4.93&4.49&3.50&3.93&{\bf 100\%}&0.942&0.934&0.930&0.896&0.912&{\bf 100\%}\\
			GCPN&7.98&7.85&7.80&-&-&{\bf 100\%}&{\bf0.948}&0.947&0.946&-&-&{\bf 100\%}\\
			MHG-VAE&5.56&5.40&5.34&4.12&4.49&{\bf 100\%}&0.947&0.946&0.944&0.920&0.929&{\bf 100\%}\\
			MSO&14.44&14.20&13.95&13.49&13.67&-&{\bf0.948}&{\bf0.948}&{\bf0.948}&{\bf0.948}&{\bf0.948}&-\\
			\midrule
			MNCE-RL$_{OEU}$&{14.49}&{14.44}&{14.36}&{14.13}&{14.16}&{\bf100\%}&{\bf0.948}&{\bf0.948}&{\bf0.948}&{\bf 0.948}&{\bf 0.948}&{\bf 100\%}\\
			MNCE-RL&{\bf18.33}&{\bf18.18}&{\bf18.16}&{\bf17.52}&{\bf17.76}&{\bf 100\%}&{\bf0.948}&{\bf0.948}&{\bf0.948}&{\bf 0.948}&{\bf 0.948}&{\bf 100\%}\\
			\bottomrule
		\end{tabular*}
	\end{table}
	
	{\bf Constrained property optimization.} This task aims at generating molecules with an improved penalized logP score while keeping structures similar to a given target molecule. Different from previous methods, such as GCPN, that can generate novel molecules starting from a given molecule, we first train our model to maximize the log-likelihood of the target molecule and then optimize the penalized logP. The task-specific reward assigns a small constant score if the similarity drops below the threshold and assigns a linear projection of the penalized logP score if the similarity is larger than the threshold. The results are shown in Table \ref{cop} and Appendix G, where the $\delta$ is the threshold of the similarity score. MNCE-RL is capable to optimize all the molecules with success rates of 100\% on both thresholds and for each threshold, MNCE-RL achieves significantly higher improvements in penalized logP than all baseline models. Although the average similarity scores of the molecules generated by MNCE-RL are slightly lower than those generated by the baseline models, the improvements in penalized logP achieved by MNCE-RL with similarity threshold 0.6 is significantly higher than baseline models with threshold 0.4, exhibiting the superiority of MNCE-RL.
	
	\begin{table}[h]
		\setlength\tabcolsep{1.9pt} 
		\caption{Results on constrained property optimizations}
		\label{cop}
		\centering
		\begin{tabular*}{0.9\textwidth}{ccccccc}
			\toprule
			\multirow{3}{*}{Method} &\multicolumn{3}{c}{$\delta=0.4$}& \multicolumn{3}{c}{$\delta=0.6$}\\
			\cmidrule(lr){2-4} \cmidrule(lr){5-7}
			& Improvement &Similarity&Success &Improvement&Similarity&Success\\
			\cmidrule(lr){1-1}\cmidrule(lr){2-4} \cmidrule(lr){5-7}
			JT-VAE&$0.84\pm 1.45$&{$ 0.51\pm 0.10$}&83.6\%&$0.21\pm0.71$&{$0.69\pm0.06$}&46.4\%\\
			GCPN&$2.49\pm1.30$&$0.47\pm0.08$&{\bf 100\%}&$0.79\pm0.63$&$0.68\pm0.08$&{\bf 100\%}\\
			MHG-VAE&$1.00\pm1.87$&$\bf 0.52\pm0.11$&43.5\%&$0.61\pm1.20$&$\bf 0.70\pm 0.06$&17.0\%\\
			\midrule
			MNCE-RL &{$\bf5.29\pm1.58$}&$0.45\pm0.05$&{\bf 100\%}&{$\bf 3.87\pm1.43$}&$0.64\pm 0.04$&{\bf 100\%}\\
			\bottomrule
		\end{tabular*}
	\end{table}
	
	{{\bf Comprehensive evaluations with GuacaMol.} These experiments comprehensively measure a model's ability in optimizing properties with unlimited evaluations. The results are shown in Table \ref{guacamole}, where BNGM represents the best results of the naive baselines provided in the manuscript of GuacaMol \cite{brown2019guacamol}. The performance of MNCE-RL exceeds the baselines on all benchmarks. In particular, our method significantly outperforms the baselines in multi-objective optimization tasks, showing the superiority of MNCE-RL in complex scenarios.}
	\begin{table}[h]
		\setlength\tabcolsep{1.4pt} 
		\caption{Results on the benchmarks provided by GuacaMol.}
		\label{guacamole}
		\centering
		\begin{tabular*}{1.0\textwidth}{cccccccc}
			\toprule
			\multirow{3}{*}{Benchmark}&\multicolumn{3}{c}{Methods}&\multirow{3}{*}{Benchmark}&\multicolumn{3}{c}{Methods}\\
			\cmidrule(lr){2-4}\cmidrule(lr){6-8}
			& BNGM  &MSO &MNCE-RL&& BNGM  &MSO &MNCE-RL\\
			\cmidrule(lr){1-4}\cmidrule(lr){5-8}
			Celecoxib rediscovery  &{\bf 1.0}&{\bf 1.0}&{\bf 1.0}&Osimertinib MPO&0.953&0.966&{\bf 1.0}\\
			Troglitazone rediscovery &{\bf 1.0}&{\bf 1.0}&{\bf 1.0}&	Fexofenadine MPO&0.998&{\bf 1.0}&{\bf 1.0}\\
			Thiothixene	rediscovery &{\bf 1.0}&{\bf 1.0}&{\bf 1.0}&Ranolazine MPO&0.920&0.931&{\bf 0.990}\\
			Aripiprazole similarity&{\bf 1.0}&{\bf 1.0}&{\bf 1.0}&Perindopril MPO&0.808&0.834&{\bf 0.882}\\
			Albuterol similarity&{\bf 1.0}&{\bf  1.0}&{\bf 1.0}&Amlodipine MPO&0.894&0.900&{\bf 0.920}\\
			Mestranol similarity&{\bf 1.0}&{\bf  1.0}&{\bf 1.0}&Sitagliptin MPO&0.891&0.868&{\bf 0.904}\\
			C11H24&0.993&0.997&{\bf 1.0}&Zaleplon MPO&0.754&0.764&{\bf 0.781}\\
			C9H10N2O2PF2Cl&0.982&{\bf 1.0}&{\bf 1.0}&Valsartan SMARTS&0.990&0.994&{\bf 1.0}\\
			Median molecules 1&0.438&0.437&{\bf 0.455}&Scaffold Hop&{\bf 1.0}&{\bf 1.0}&{\bf 1.0}\\
			Median molecules 2&0.432&0.395&{\bf 0.457}&Deco Hop&{\bf 1.0}&{\bf 1.0}&{\bf 1.0}\\
			\bottomrule
		\end{tabular*}
	\end{table}

	{\bf Property range targeting.} This experiment measures the model's ability to generate diverse molecules with some specific property in a predefined range \cite{you2018graph}, where the diversity is defined as the average pairwise Tanimoto distance between the Morgan fingerprints of the generated molecules \cite{rogers2010extended}. Penalized logP and molecular weight (MW) are considered in this task where the predefined ranges are the same as those used in \cite{you2018graph}. The task-specific reward in our approach is inversely proportional to the distance between the property score of a generated molecule and the center of the predefined range.
	The results are shown in Table \ref{target}. Our model achieves over 90\% success rates in all the four tasks with high diversities \cite{you2018graph} and an over 99\% success rate in targeting the range $500\leq MW\leq 550$, which significantly outperforms state-of-the-art methods. 
	
	\begin{table}[h]
		\setlength\tabcolsep{1.0pt} 
		\caption{Results on property range targeting}
		\label{target}
		\centering
		\begin{tabular*}{0.95\textwidth}{ccccccccc}
			\toprule
			\multirow{3}{*}{Method} &\multicolumn{2}{c}{$-2.5\leq logP\leq -2$}& \multicolumn{2}{c}{$5\leq logP\leq 5.5$}&\multicolumn{2}{c}{$150\leq MW\leq 200$}&\multicolumn{2}{c}{$500\leq MW \leq550$}\\
			\cmidrule(lr){2-3} \cmidrule(lr){4-5}  \cmidrule(lr){6-7}  \cmidrule(lr){8-9}
			& Success &Diversity &Success&Diversity&Success &Diversity&Success &Diversity\\
			\cmidrule(lr){1-1}\cmidrule(lr){2-3} \cmidrule(lr){4-5}  \cmidrule(lr){6-7}  \cmidrule(lr){8-9}
			JT-VAE&11.3\%&{\bf 0.846}&7.6\%&{\bf 0.907}&0.7\%&0.824&16.0\%&0.898\\
			GCPN&85.5\%&0.392&54.7\%&0.855&76.1\%&0.921&74.1\%&{\bf0.920}\\
			\midrule
			MNCE-RL &{\bf98.3\%}&0.836&{\bf 98.0\%}&0.842&{\bf91.8\%}&{\bf 0.928}&{\bf 99.6\%}&0.870\\
			\bottomrule
		\end{tabular*}
	\end{table}

	{\bf Property optimization with limited property evaluations.} This task measures a model's ability to optimize molecules when the property evaluation is expensive. As done in \cite{kajino2019molecular}, we limit the number of molecule property queries to 500. We repeat MNCE-RL ten times and take the first 500 molecules generated as the output each time to obtain a total of 5k molecules. The task-specified reward is the same as in property optimization with unlimited property evaluations. The top 3 property scores, the 50th best score and the average score of the top 50 molecules are recorded. The results are shown in Table \ref{lpop} and Appendix G. Our model significantly outperforms all baseline methods. Interestingly, even with limited property evaluations, our method still performs better than JT-VAE and MHG-VAE with unlimited evaluations. Moreover, the top 50 scored molecules generated by MNCE-RL has a higher average penalized logP score than the top-scored molecule generated by all baselines, which demonstrates the superiority of MNCE-RL in situations when it is expensive to evaluate molecular properties.
	
	\begin{table}[h]
		\setlength\tabcolsep{1.9pt} 
		\caption{Results on property optimization with limited property evaluations}
		\label{lpop}
		\centering
		\begin{tabular*}{0.52\textwidth}{ccccccc}
			\toprule
			\multirow{3}{*}{Method} &\multicolumn{6}{c}{Penalized logP}\\
			\cmidrule(lr){2-7}
			& $1^{st}$ &$2^{nd}$ &$3^{rd}$&$50^{th}$&\vtop{\hbox{\strut Top 50}\hbox{\strut\ \ Avg.}} &Validity\\
			\cmidrule(lr){1-1}\cmidrule(lr){2-7}
			JT-VAE&1.69&1.68&1.60&-9.93&-1.33&{\bf100\%}\\
			GCPN&2.77&2.73&2.34&0.91&1.36&{\bf100\%}\\
			MHG-VAE&5.24&5.06&4.91&4.25&4.53&{\bf100\%}\\
			MSO&2.96&2.91&2.75&2.49&2.54&{\bf100\%}\\
			\midrule
			MNCE-RL &{\bf9.88}&{\bf9.82}&{\bf9.75}&{\bf7.28}&{\bf8.31}&{\bf100\%}\\
			\bottomrule
		\end{tabular*}
	\end{table}
	
	{\bf Generation of novel molecules with antibacterial property.} This experiment shows MNCE-RL's ability to assist drug discovery in a real-world application scenario when the number of experimentally validated molecules is limited and there is no known evaluation function. We first train a classifier on the 2,337 molecules from \cite{stokes2020deep} to distinguish positive and negative samples and use the classifier as a pseudo evaluation function. Then, we extract production rules from these molecules. The problem is modeled as a property optimization where we try to find molecules that receive high scores from the classifier. As the classifier is severely overfitted, when training the generation model, we assume that the generated novel molecules are negative and use these "negative samples" to update the classifier to reduce bias. After training, the kinase inhibitor scores, the protease inhibitor scores, and the enzyme inhibitor scores \cite{schaenzer2017screen, umezawa1982low,el2016synthesis, alodeani2015anti} (see Appendix E for details) of the top 10 molecules with the highest scores assigned by the classifier are reported. The results are shown in Appendix G and Table \ref{anti}. Ten of the ten molecules are bioactive (with scores larger than 0.2; see Table \ref{anti}) with at least one inhibitor score, and six of them are highly bioactive (with scores larger than 0.5), which illustrates the ability of MNCE-RL to generate antibacterial candidate molecules with only limited labeled samples.
	\begin{table}[h]
		\setlength\tabcolsep{1.9pt} 
		\caption{Properties of six molecules having high inhibitor scores.}
		\label{anti}
		\centering
		\begin{tabular*}{0.8\textwidth}{cccc}
			\toprule
			\multirow{3}{*}{Molecule}&\multicolumn{3}{c}{Computed properties}\\
			\cmidrule(lr){2-4}
			& Kinase inhibitor (KI) &Protease inhibitor (PI) &Enzyme inhibitor (EI)\\
			\cmidrule(lr){1-1}\cmidrule(lr){2-4}
			$M_1$&-0.38&0.56&0.25\\
			$M_2$&-0.20&0.54&0.23\\
			$M_3$&-0.34&0.55&0.15\\
			$M_4$&-0.24&0.63&0.09\\
			$M_5$&-0.16&0.66&0.16\\
			$M_6$&-0.24&0.66&0.30\\
			\bottomrule
		\end{tabular*}
	\end{table}

	\section{Conclusion and future work}
	In this paper, we propose a new method MNCE-RL based on the novel molecular NCE grammars to solve the molecular optimization problem in the RL framework. MNCE-RL achieves the state-of-the-art performance in a series of systematic experiments. In a real-world application, when the molecules with known properties are limited and no numerical evaluation function is known, our method still exhibits high potential to generate molecules with desired properties, showing its great potential utility in drug discovery. {Although our proposed grammar guarantees the valency validity of the generated structures, it struggles to capture high-level chemical properties such as bond orders. We leave it to future work.}

	\section*{Broader impact}
	Finding effective medicines for diseases has always been a challenge in the pharmaceutical industry, especially when precision medicine has attracted more and more attention in recent years. Our approach provides an efficient way to generate molecules with specific properties, which will help reduce the workload of pharmacists, accelerate the development of novel drugs, and decrease the cost of drug design. On the other hand, although the molecules generated by our method possess desirable biological or chemical properties, their safety and effectiveness on patients still need to be validated in the normal clinical trial processes.
	
	\begin{ack}
	This work has been supported in part by the National Natural Science Foundation of China grant
	61772197, the National Key Research and Development Program of China grant
	2018YFC0910404 and the Guoqiang Institute of Tsinghua University with grant no. 2019GQG1.
	\end{ack}
	\bibliographystyle{abbrv}
	\bibliography{reference}

\end{document}


\counterwithin{figure}{section}
	\begin{appendices}
		\section{Supplementary figures}
		\label{figures}
		
		\begin{figure}[h]
			\centering
			\includegraphics[width=\textwidth]{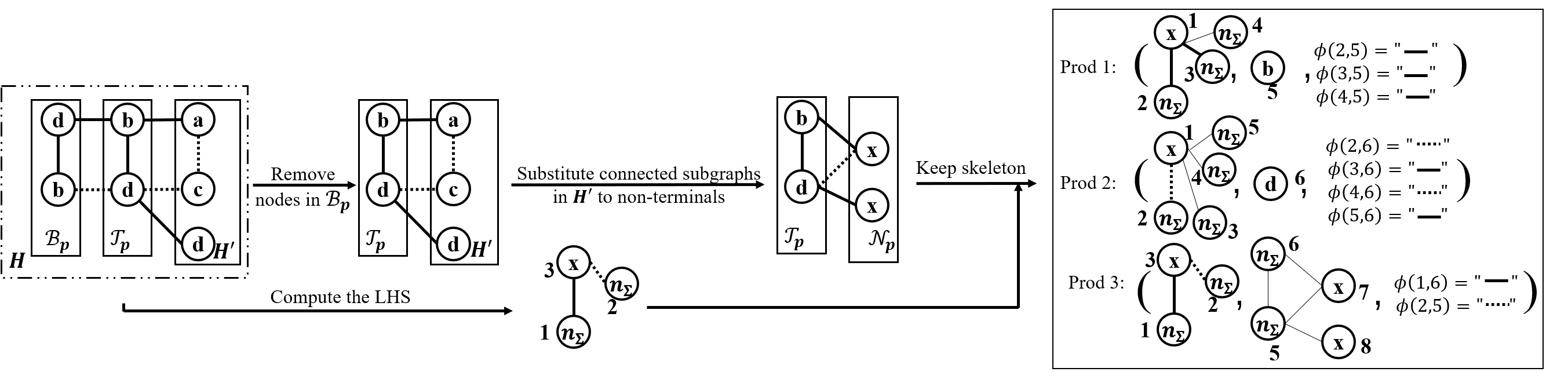}
			\caption{Extraction of complex production rules. The LHS is computed by representing the nodes in $\mathcal{T}_p$ as a non-terminal node, removing the edges between nodes in $\mathcal{B}_p$ and labeling the nodes in $\mathcal{B}_p$ as $n_\Sigma$. The computation of RHS is removing the nodes in $\mathcal{B}_p$ and turning the connected subgraphs in $H'$ to non-terminal nodes. To reduce the number of production rules, only the skeleton of the RHS is kept and a production rule for each node in $\mathcal{T}_p$ is introduced to maintain the information.}
			\label{production1}
		\end{figure}
		\begin{figure}[h]
			\centering
			\includegraphics[width=\textwidth]{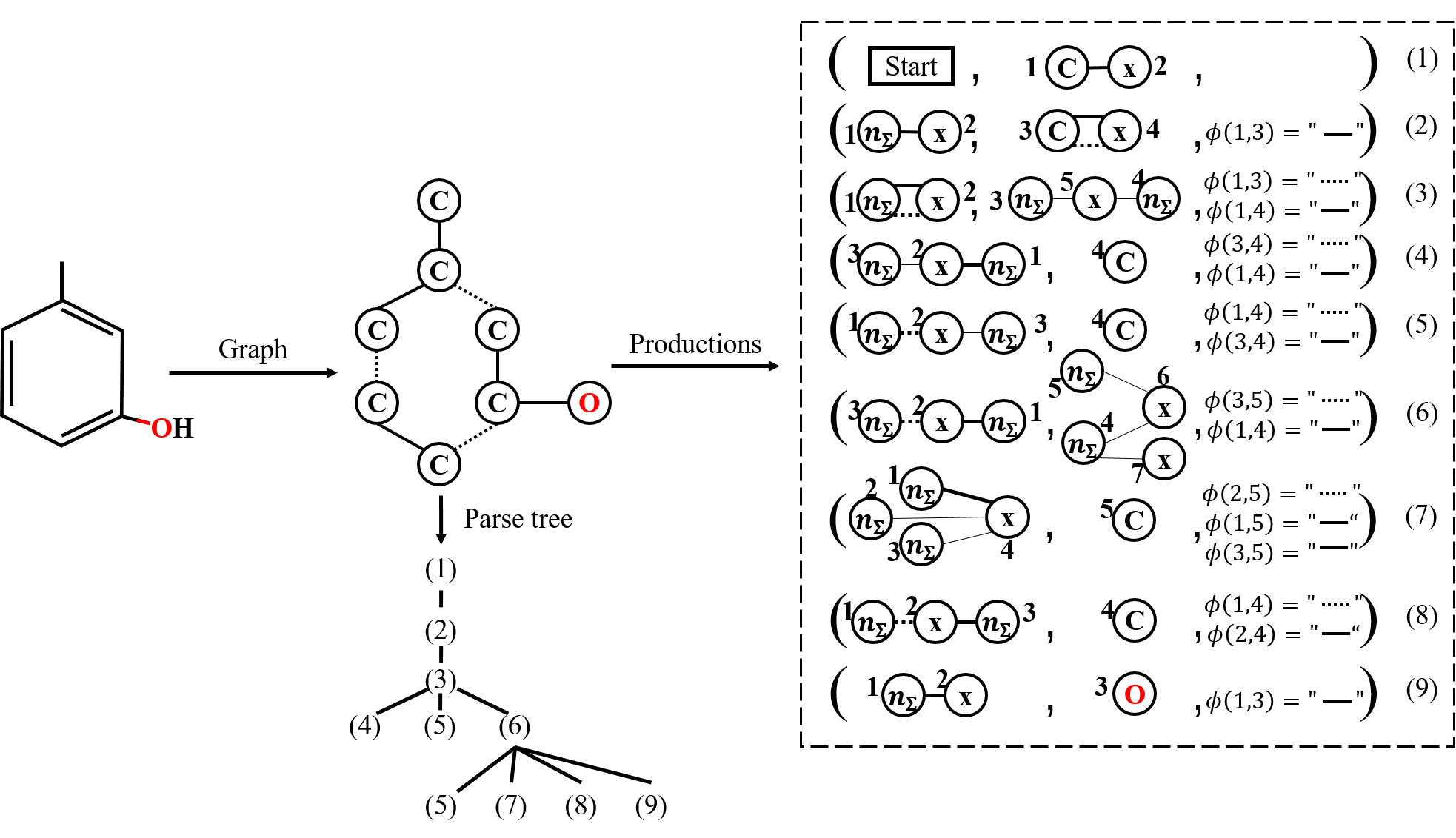}
			\caption{An example of transforming a molecule into a parse tree and inferring molecular NCE grammar production rules.}
			\label{example1}
		\end{figure}
		
		\begin{figure}
			\centering
			\includegraphics[width=\textwidth]{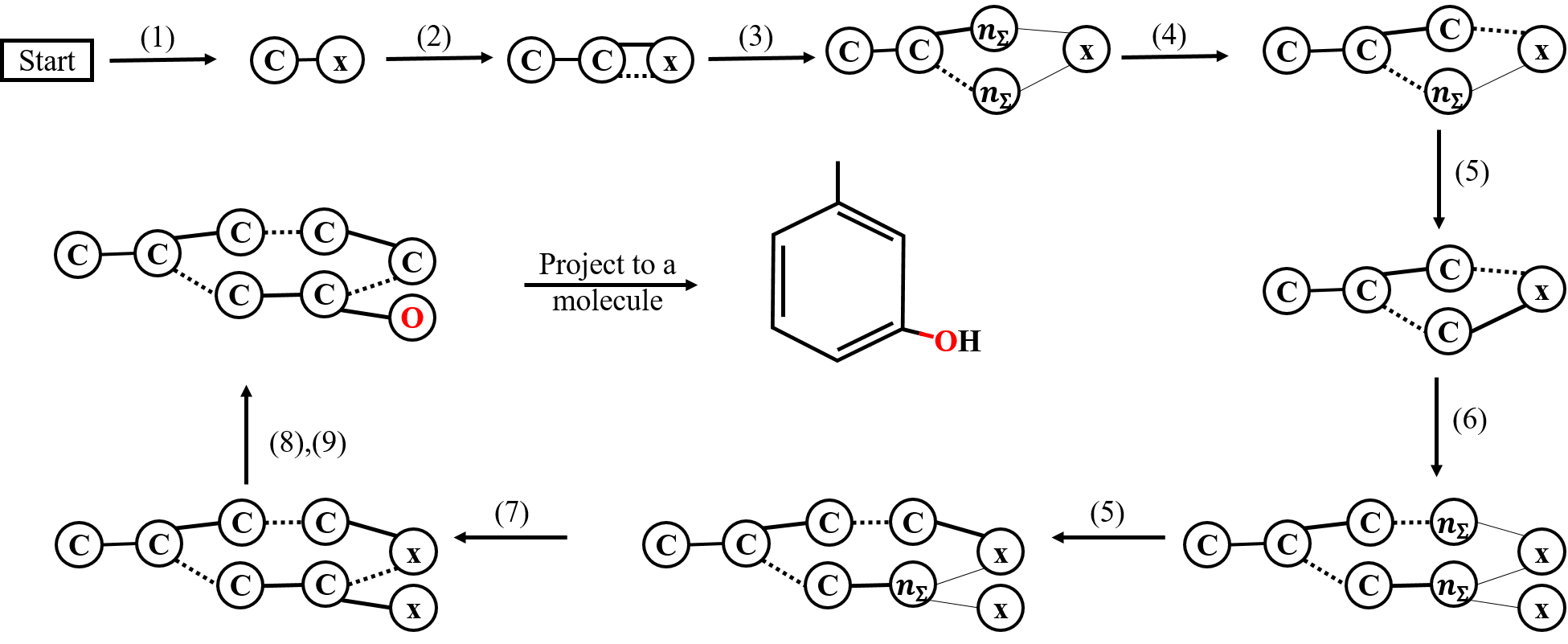}
			\caption{An example of sampling a molecule from a molecular NCE grammar. The production rules are shown in Figure \ref{example1}.}
			\label{example2}
		\end{figure}

		\newpage
		\section{Supplementary information of the proposed grammars}
		\label{algopi}
		The algorithm to infer production rules of the molecular NCE grammar and parse molecular graphs into parse trees is shown in Algorithm \ref{Parsegraph}, where $v_T$ is a node of the parse tree $T$ and $Neigh(v)_H$ is the set of first-hop neighbors of a node $v$ in the graph $H$. For a molecular graph $H$ with $\|V_H\|$ nodes, the time complexity of Algorithm \ref{Parsegraph} is $O(\|V_H\|^2)$. 
		
		{Compared with MHG, our proposed grammars have better generalization ability. MHG is an extension of hyperedge replacement grammar, which is based on the clique tree decomposition of graphs. In molecular hypergraphs, the clique tree decomposition might introduce a large number of rare substructures and cause a low coverage rate. For instance, in the MHG inferred from the ZINC250k dataset, 1,424/2,031 are starting rules and 2/3 of the starting rules are used by less than ten molecules. At the same time, 16/5,000 molecules in the testing set cannot be covered by these inferred production rules. In comparison, our grammar is based on neighboring relationships. In molecular graphs, the degree and neighbors of each node are limited by chemical rules, thus the substructure involved in our grammar is relatively simpler and in smaller fragments, which leads to fewer production rules and a higher coverage rate (see Appendix \ref{bsmg}).}
		
		{In the generation process, to be consistent with the inference process, the non-terminals with labels of $n_\Sigma$ have higher priority than non-terminals with labels of $x$, and the non-terminals that are generated later have higher priority. The non-terminal with the highest priority in the intermediate graph is rewritten each time.}
		\begin{algorithm}
			\label{Parsegraph}
			\SetAlgoLined
			\KwIn{$H$, $P$, $\mathcal{B}_p$, $\mathcal{T}_p$, $v_T$}
			\KwOut{$T$, $P$}
			\SetKwFunction{FMain}{ParseMolecularGraph}
			\SetKwProg{Fn}{Function}{:}{}
			\Fn{\FMain{$H$, $P$, $\mathcal{B}_p$, $\mathcal{T}_p$, $v_T$}}
			{
				\If{$\mathcal{B}_p$ is empty}
				{	Initialize tree $T$\;
					Initialize $v_T$ as the root of $T$\;
					Add the initial node to $\mathcal{B}_p$\;
					Arbitrarily select a node from $H$ and add it to $\mathcal{T}_{p}$\;
				}
				
				Compute $LHS$\;
				Record the embedding function $\phi$\;
				Denote $H'$ as a node-induced subgraph of $H$ where $V_{H'}=V_H\setminus\left(\mathcal{B}_p\cup\mathcal{T}_p\right)$\;	
				Remove nodes in $\mathcal{B}_p$ from $H$\;
				Represent connected subgraphs in $H'$ by non-terminal nodes\;
				Obtain the $RHS$\;
				\If{$\|\mathcal{T}_p>1\|$}
				{
					\For{$v$ in $\mathcal{T}_p$}
					{
						Add a child node $v_c$ to $v_T$\;
						Extract a production rule $p_v$ for $v$\;
						Label $v_c$ as $p_v$\;
						Add $p_v$ to $P$\;
					}
					$RHS\longleftarrow$The skeleton of $RHS$\;
					
				}
				
				$p\longleftarrow\left(LHS, RHS, \phi \right)$\;
				Label $v_T$ as $p$\;
				
				Add $p$ to $P$\;
				$\mathcal{T}^{(descent)}\longleftarrow \cup_{v\in\mathcal{T}_p}Neigh(v)_H$\;
				\For{connected subgraph $h$ in $H'$}
				{
					Add a child node $v_c$ to $v_T$\;
					$\mathcal{B}^{(h)}\longleftarrow \left(\cup_{v\in V_{h}}Neigh(v)_H\right)\setminus V_{h}$\;
					$\mathcal{B}_{p}^{(h)}\longleftarrow \mathcal{T}_{p}\cap \mathcal{B}^{(h)}$\;
					$\mathcal{T}_{p}^{(h)}\longleftarrow \mathcal{T}^{(descent)}\cap V_{h}$\;
					Denote $H^{(h)}$ as an induced subgraph of $H$, where $V_{H^{(h)}}=V_h\cup \mathcal{B}_{p}^{(h)}$\;
					\FMain{$H^{(h)}$, $P$, $\mathcal{B}_{p}^{(h)}$, $\mathcal{T}_{p}^{(h)}$, $v_c$}\;
				}
				\textbf{return} $T,P$\;
			}
			\textbf{End Function}
			\caption{Inference of molecular NCE grammar production rules}
		\end{algorithm}
		\newpage
		\section{Basic statistics of the inferred molecular NCE grammars}
		\label{bsmg}
		First, we report the basic statistics of the molecular NCE grammars inferred from the ZINC250k dataset. 
		
		To check the generalization ability of the molecular NCE grammars, we parsed the molecules in the training data. From the 220,011 training molecules, we obtained 1,775 production rules. To investigate the coverage rate of the grammar, we parsed the 5,000 molecules in the test data using the production rules inferred from the training data to estimate the percentage of molecules that cannot be represented by the inferred grammar. The result shows that only 3 out of the 5,000 molecules cannot be parsed. Our coverage rate is higher than the one achieved by the MHGs and the number of our production rules is less. 
		
		Next, we inferred grammatical production rules from all 250k ZINC250k molecules, resulting in 1,838 production rules in total. Each molecule is associated with 28 production rules on the average. The maximum number of production rules associated with a molecule is 51.
		
		For the antibiotic dataset, we extracted production rules from all known molecules. We parsed the molecules starting from different nodes to extend the number of training production sequences. 3,897 production rules were obtained from the dataset.
		
		For the data provided by GuacaMol, 7256 production rules were obtained from the training set, leading to 293/238708 molecules in the test set uncovered. In comparison, 13110 production rules were obtained for the MHGs and 1088/238708 molecules were not covered by the inferred MHG.
		
		When training the generation model, we set $L_{max}$ as the maximum number of production rules that a molecule in the dataset may be associated with. During the test, we set $L_{max}$ as $\infty$ in the experiments on ZINC250k and GuacaMol, but in the antibacterial experiment, considering the application scope of the classifier, we set $L_{max}$ the same as in the training process.
		
		\section{Experimental settings of the baseline methods}
		\label{baseline}
		Four state-of-the-art methods are compared with our method. 1) Junction tree VAE (JT-VAE) is a state-of-the-art algorithm for generating molecular graphs under the VAE framework. The basic idea of JT-VAE is to generate molecular graphs cluster by cluster and join each generated cluster using a greedy search. JT-VAE can generate molecules with 100\% validity and it outperformed the previous methods such as Syntax-directed VAE and grammar VAE in property optimization and constrained property optimization. 2) Graph convolutional policy network (GCPN) aims to generate molecules atom by atom and optimize the properties of molecules by RL. As chemical validity cannot be guaranteed intrinsically in GCPN, it checks the validity of the graph in each step and discards invalid parts. A beam search is used in GCPN to improve sampling efficiency. GCPN achieved much better performance in property optimization, property range targeting and constrained optimization than the previous methods including JT-VAE. 3) Molecular hypergraph grammar variational autoencoder (MHG-VAE) uses molecular hypergraph grammars (MHGs) to assist the generation of molecular graphs and focuses on generating molecules with limited property evaluations. MHG-VAE uses an MHG as the prior of its VAE model, and achieved better performance than GCPN and JT-VAE under the limited property evaluation setting, but showed no advantage over other methods when property evaluation was unlimited. 4) Molecule Swarm Optimization (MSO) is a state-of-the-art algorithm in multi-objective molecular optimization with the particle swarm optimization algorithm and achieved excellent performance on the benchmarks provided by GuacaMol. The codes of the baselines were downloaded from \href{https://github.com/bowenliu16/rl_graph_generation}{GCPN}, \href{https://github.com/wengong-jin/icml18-jtnn}{JT-VAE}, \href{https://github.com/ibm-research-tokyo/graph_grammar} {MHG-VAE} and \href{https://github.com/jrwnter/mso}{MSO}.
		
		{\bf Property optimization with unlimited property evaluations.} The results of GCPN were copied from \cite{you2018graph}. As JT-VAE provided the molecules it generated when optimizing the penalized logP, we obtained the results directly by scoring the provided molecules. As for the task of optimizing QED, we set the objective function as the QED score and ran the code of JT-VAE with the default setting ten times to generate novel molecules. The results were obtained by summarizing all the molecules generated in the ten runs. For MHG-VAE, we copied its results in optimizing penalized logP from \cite{kajino2019molecular} and obtained the results in optimizing QED by running its code in the default setting with the QED score as the objective function. For MSO, to fairly compare with our method, the results in Table 1 were obtained by constraining the maximum number of atoms to 51 and the best hyperparameters used in the corresponding paper \cite{winter2019efficient} were adopted in our experiments. We ran MSO 100 times and merge all the obtained molecules as the results. As a comparison, the results of MSO without constraints on the number of atoms as well the results of our method under relaxed constraints are shown in Table \ref{ablation}.
		
		{\bf Constrained property optimization.} The results of all baselines were copied from the corresponding papers \cite{you2018graph,kajino2019molecular,jin2018junction}.
		
		{\bf Comprehensive evaluations with GuacaMol.} The results of all baselines were copied from the corresponding papers \cite{winter2019efficient, brown2019guacamol}.
		
		{\bf Property range targeting.} The results of GCPN and JT-VAE were directly copied from \cite{you2018graph}.

		{\bf Property optimization with limited evaluations:} The results of JT-VAE and GCPN were copied from \cite{kajino2019molecular}. For MHG-VAE, we ran the code ten times and took the first 250 molecules each time, with the same hyperparameters used in \cite{kajino2019molecular}. For MSO, we ran their code ten times and took the first 500 molecules each time, with the default hyperparameters.
		
		\section{Evaluation of antibacterial properties}
		\label{ins}
		Enzymes are biological catalysts. A protease is an enzyme that performs proteolysis, that is, it triggers protein catabolism by hydrolysis of the peptide bonds that link amino acids together in a polypeptide chain. A kinase is an enzyme that catalyzes the transfer of phosphate groups from high-energy, phosphate-donating molecules to specific substrates. As enzymes play an important role in bacterial activities, molecules with high enzyme inhibitor scores, protease inhibitor scores or kinase inhibitor scores are thought to be high-potential candidates for antibiotics.
		
		The inhibitor scores were computed by using the \href{https://www.molinspiration.com/cgi-bin/properties}{Molinspiration online server}. The larger the score is, the higher is the probability that the involved molecule will be active. In particular, molecules with positive scores are usually thought to be active. In our experiment, we adopted the thresholds used by Molinspiration and regarded those with scores larger than 0.2 as active molecules and those with scores larger than 0.5 as highly active molecules.
		
		\newpage
		\section{Supplementary details of model training}
		\label{ModelTrain}
		The model is pre-trained with known molecules by maximizing the likelihood and then trained for each optimization task. The hyperparameters in reward functions are optimized for each task independently. For tasks with unlimited property evaluations, the other hyperparameters are optimized on the optimizing penalized logP task. The hyperparameters for the optimization with limited property evaluations are optimized independently. For each task, the best molecules found by the policy are used as known trajectories to train the model to accelerate convergence. With 1080Ti, the pre-training on ZINK250 took around 27 hours and the optimization stages took 30 minutes $\sim$ 24 hours depending on the tasks. 
		
		\newpage
		
		\section{Supplementary results}
		\label{sresult}
		\begin{figure}[h]
			\centering
			\includegraphics[width=\textwidth]{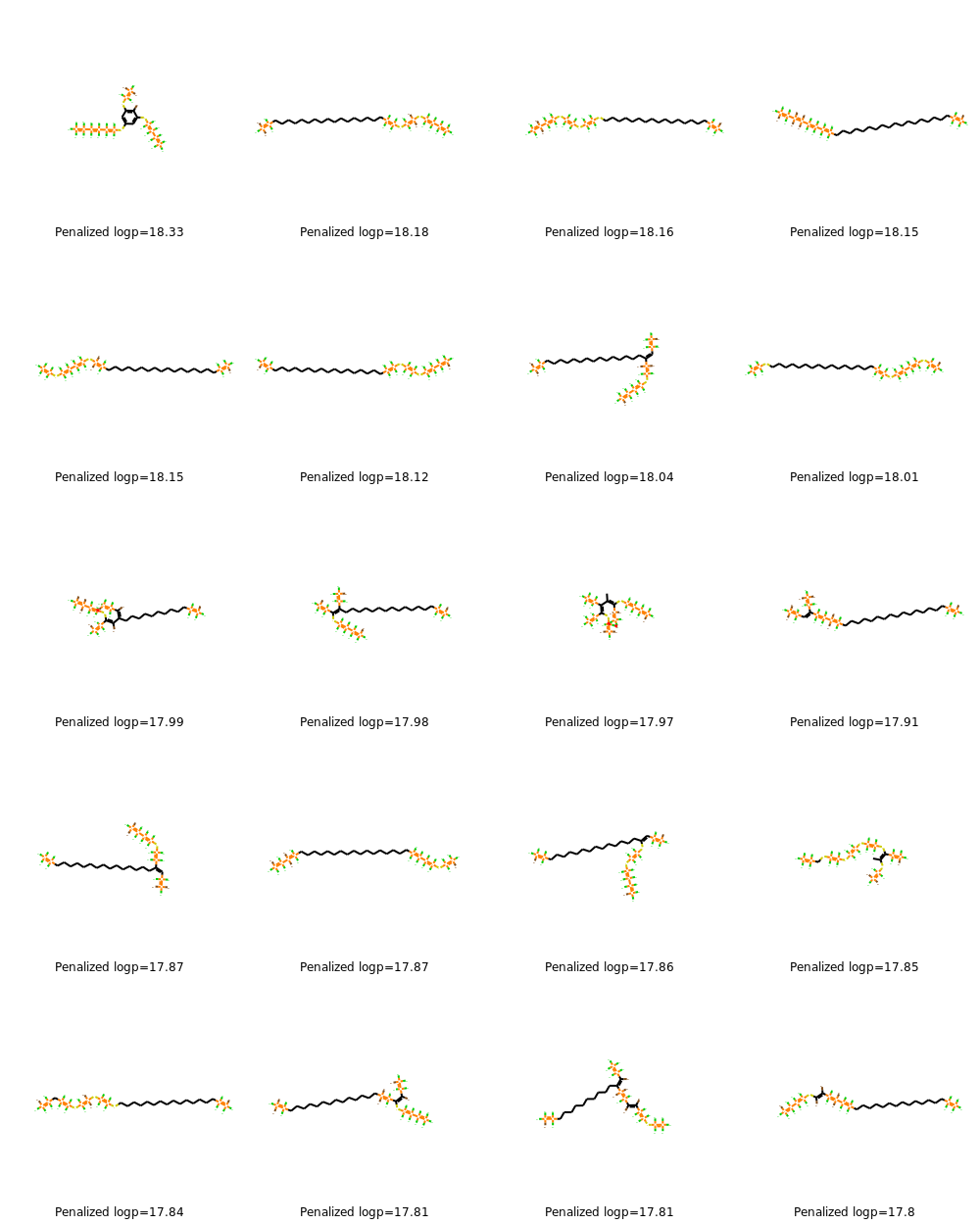}
			\caption{The 20 molecules with the highest penalized logP scores generated by MNCE-RL in optimizing the penalized logP score with unlimited property evaluations. The diversity of 5000 molecules is 0.722.}
			\label{optlogp}
		\end{figure}
		
		\begin{figure}[b]
			\centering
			\includegraphics[width=1.0\textwidth]{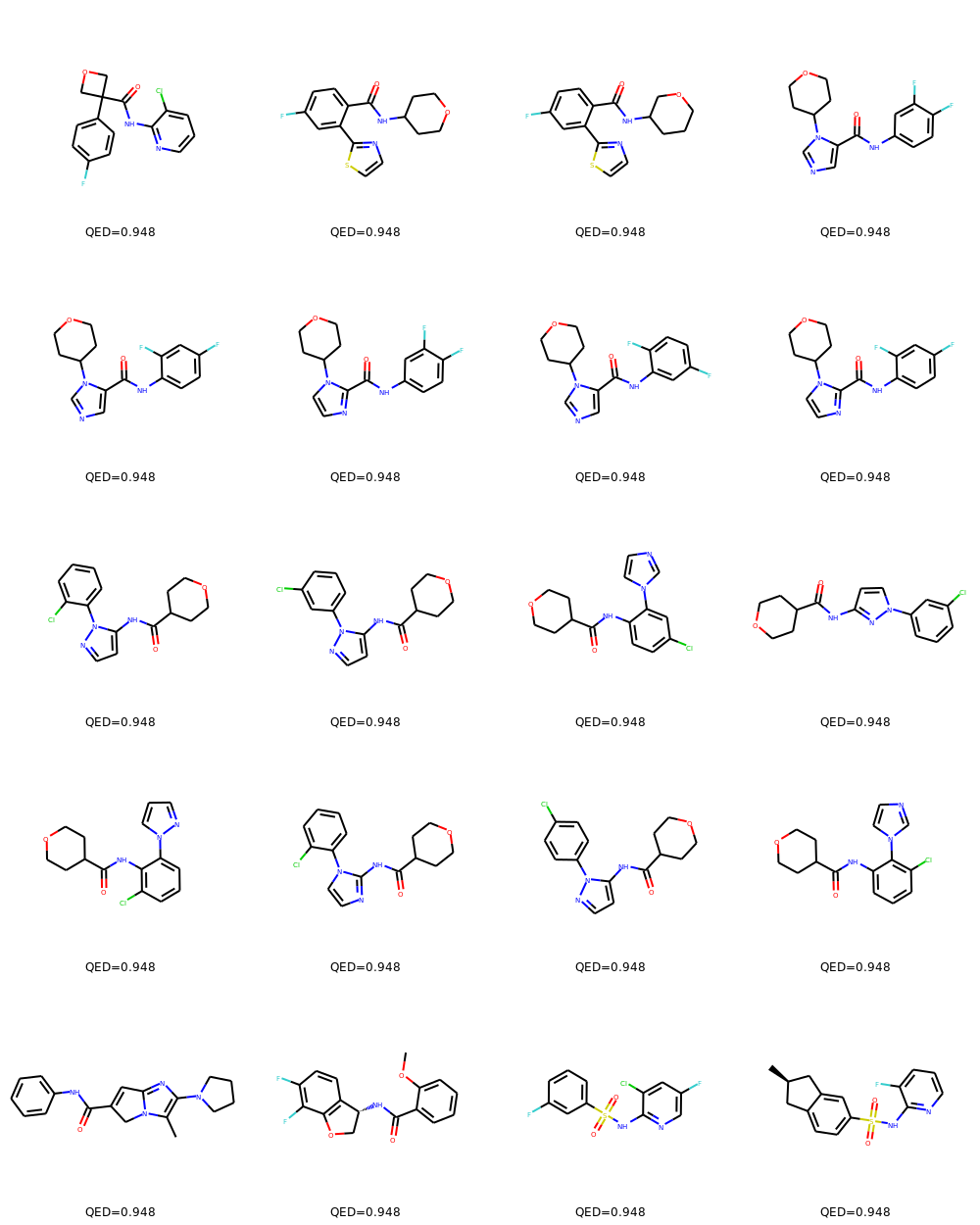}
			\caption{The 20 molecules with the highest QED scores generated by MNCE-RL in optimizing the QED score with unlimited property evaluations. The diversity of 5000 molecules is 0.870.}
			\label{optqed}
		\end{figure}
		
		\begin{figure}[h]
			\centering
			\includegraphics[width=0.95\textwidth]{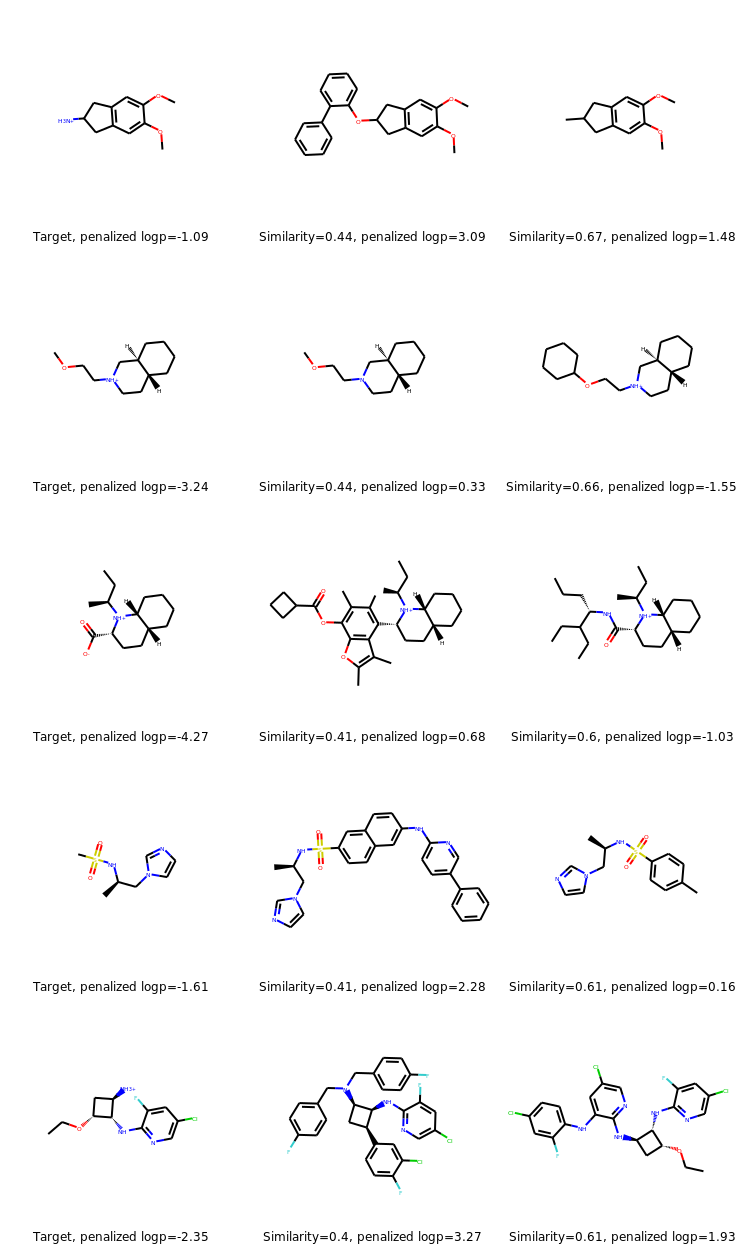}
			\caption{Five target molecules (the first column) in constrained optimization and their corresponding optimized molecules generated by MNCE-RL (the second and the third columns).}
			\label{constrain}
		\end{figure}
		\begin{figure}[h]
			\centering
			\includegraphics[width=0.75\textwidth]{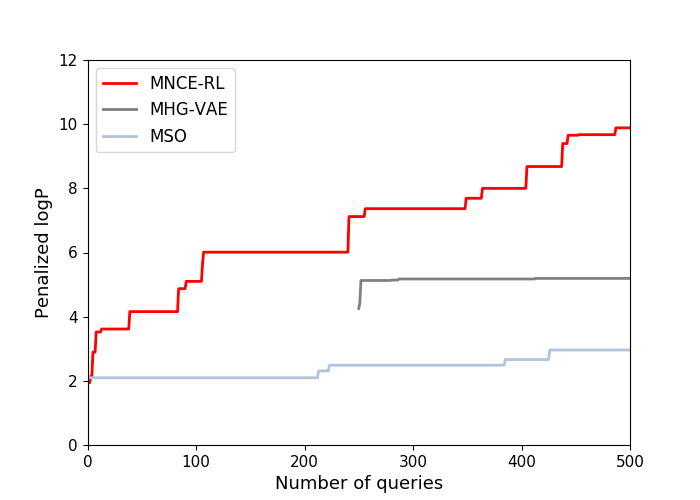}
			\caption{The best penalized logP scores of the molecules found by different methods depending on the number of function evaluations.}
			\label{limited}
		\end{figure}
		\begin{figure}[h]
			\centering
			\includegraphics[width=\textwidth]{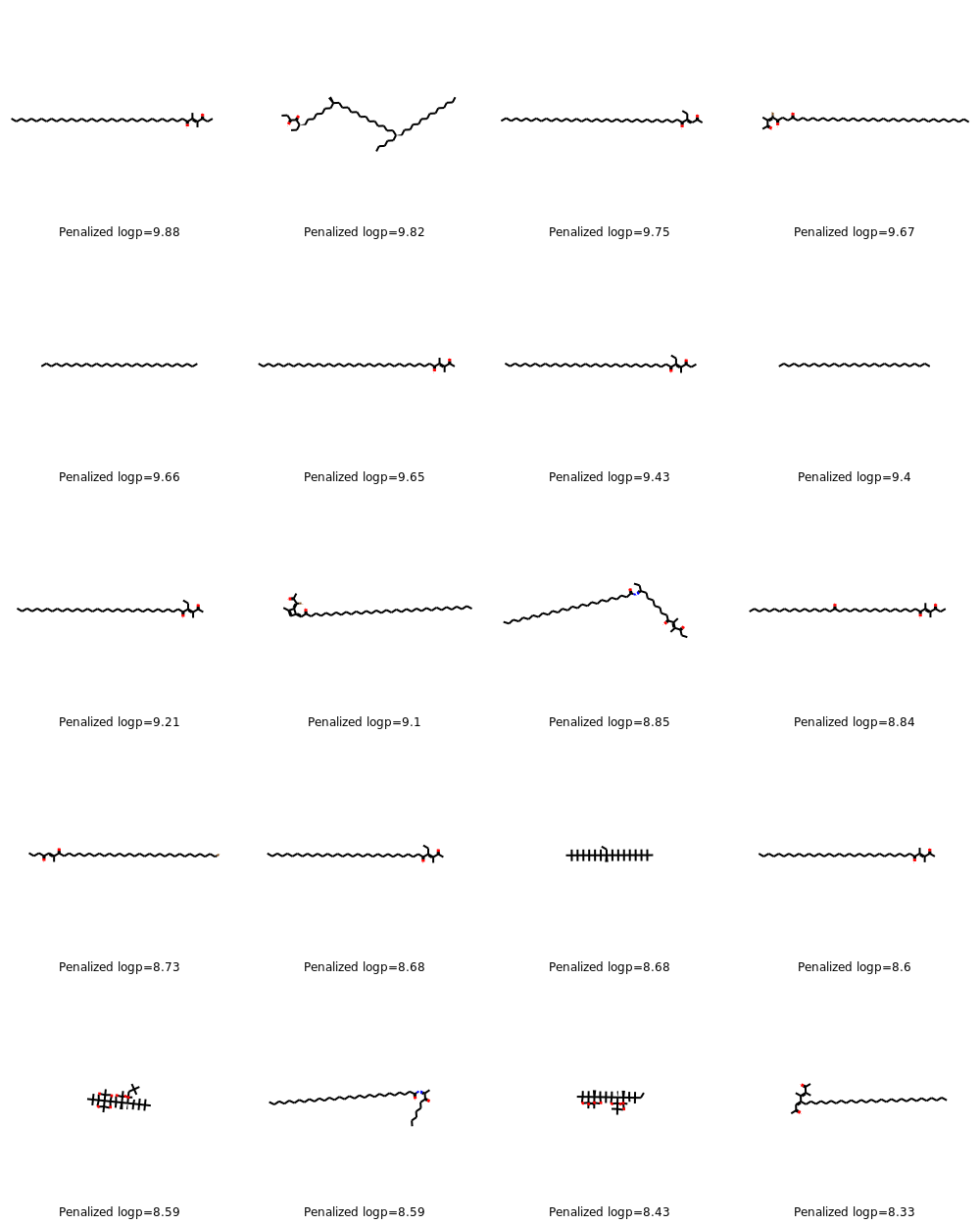}
			\caption{The 20 molecules with the highest penalized logP scores generated by MNCE-RL in optimizing the penalized logP score with limited property evaluations.}
			\label{optloplimited}
		\end{figure}
		
		\begin{figure}[h]
			\centering
			\includegraphics[width=\textwidth]{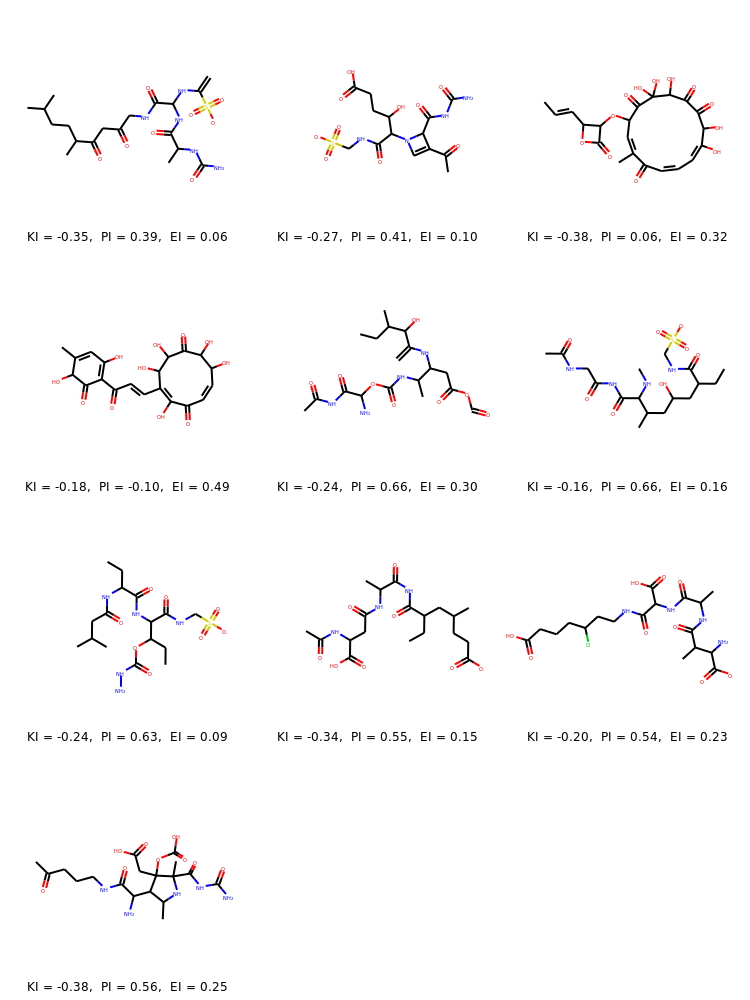}
			\caption{The ten molecules with the highest scores assigned by the classifier in generating candidates of antibiotics and their corresponding property scores.}
			\label{antibio}
		\end{figure}
		\begin{table}[h]
			\setlength\tabcolsep{1.9pt} 
			\caption{The maximum penalized logP scores with different $L_{max}$ values. The results of MSO are copied from the corresponding paper \cite{winter2019efficient}.}
			\label{ablation}
			\centering
			\begin{tabular*}{0.9\textwidth}{ccccccc}
				\toprule
				\multirow{3}{*}{Method} &\multicolumn{6}{c}{Penalized logP}\\
				\cmidrule(lr){2-7}
				& $1^{st}$ &$2^{nd}$ &$3^{rd}$&$50^{th}$&\vtop{\hbox{\strut Top 50}\hbox{\strut\ \ Avg.}} &Validity\\
				\cmidrule(lr){1-1}\cmidrule(lr){2-7}
				MSO (no constraints on the number of atoms)&26.10&-&-&-&-&-\\
				\midrule
				MNCE-RL ($L_{max}=51$) &18.33&18.18&18.16&17.52&17.76&{\bf 100\%}\\
				MNCE-RL ($L_{max}=90$) &{ 28.09}&{28.04}&{28.00}&{26.52}&{26.99}&{\bf 100\%}\\
				MNCE-RL ($L_{max}=110$) &{\bf34.06}&{\bf34.04}&{\bf33.92}&{\bf32.96}&{\bf33.33}&{\bf100\%}\\
				\bottomrule
			\end{tabular*}
		\end{table}
		
	\end{appendices}
	
	\newpage
	\bibliographystyle{abbrv}
	\bibliography{reference}